# Comparative Evaluation of Traditional and Deep Learning-Based Segmentation Methods for Spoil Pile Delineation Using UAV Images

Sureka Thiruchittampalam, Bikram P. Banerjee, Nancy F. Glenn, Simit Raval

*Abstract*— The stability of mine dumps is contingent upon the precise arrangement of spoil piles, taking into account their geological and geotechnical attributes. Yet, on-site characterisation of individual piles poses a formidable challenge. The utilisation of image-based techniques for spoil pile characterisation, employing remotely acquired data through unmanned aerial systems, is a promising complementary solution. Image processing, such as object-based classification and feature extraction, are dependent upon effective segmentation. This study refines and juxtaposes various segmentation approaches, specifically colour-based and morphology-based techniques. The objective is to enhance and evaluate avenues for object-based analysis for spoil characterisation within the context of mining environments. Furthermore, a comparative analysis is conducted between conventional segmentation approaches and those rooted in deep learning methodologies. Among the diverse segmentation approaches evaluated, the morphology-based deep learning segmentation approach, Segment Anything Model (SAM), exhibited superior performance in comparison to other approaches. This outcome underscores the efficacy of incorporating advanced morphological and deep learning techniques for accurate and efficient spoil pile characterisation. The findings of this study contribute valuable insights to the optimisation of segmentation strategies, thereby advancing the application of image-based techniques for the characterisation of spoil piles in mining environments.

## I. INTRODUCTION

THE complexity inherent in the composition of coal spoils presents a formidable challenge in the endeavour to devise stable dump structures [1]. Formulation of such structures is the initial phase of spoil pile characterisation, constituting the foundational elements of dumps. However, the on-site characterisation of individual spoil piles is encumbered by challenges, encompassing strenuous efforts, safety hazards, and susceptibility to subjective human biases. Employing a remote sensing-based approach may reduce the risks and limitations associated with on-site waste characterisation.

In recent advancements, the utilisation of a classification framework based on remote data gathering through miniaturised sensors attached to unmanned aerial vehicles (UAVs) has surfaced as an effective solution in the mining context [2-4]. This methodology not only attests to an elevated standard of safety but also affords a cost-effective and temporally efficient alternative to conventional in-field characterisation protocols. Such an approach denotes a paradigm shift towards safer, more efficient, and objective spoil pile characterisation.

Image-based characterisation of spoil piles presents unique challenges due to the inherent variability in their topography and the dynamic effects of shadows and sun angles during image acquisition [5]. These factors can lead to misclassification errors, resulting in pixels belonging to the same class being assigned to different classes or pixels from different classes being grouped together [6]. Object-based classification approaches address these issues by grouping pixels into objects and assigning each object to a specific class based on its spectral and textural characteristics. This approach provides a more realistic and geometrically accurate representation of spatial features compared to traditional pixel-based classification methods. However, segmenting high-resolution remote sensing images of complex terrain like mining spoil piles remains a challenging task. The development of robust segmentation algorithms that can consistently perform well across all objects and locations in the image is hindered by the variability in acquisition parameters, such as illumination and sharpness. Additionally, segmentation quality is inherently subjective and heavily influenced by the user's definition of objects of interest and the granularity of those objects. Consequently, segmented images often exhibit two common flaws: over-segmentation, where objects of interest are fragmented into multiple pieces, and under-segmentation, where objects of interest are merged with their surroundings [7]. Segmentation is therefore a critical step in object-based image analysis, as the accuracy of classification directly depends on segmentation quality. Given the lack of studies specifically exploring segmentation algorithms for spoil pile delineation [8], evaluating the suitability of different algorithms

This work was supported in part by the Australian Coal Industry's Research Program (ACARP) C29048. (Corresponding author: Simit Raval).

Sureka Thiruchittampalam is with the School of Minerals and Energy Resources Engineering, University of New South Wales, Sydney, Australia NSW 2052 (e-mail: s.thiruchittampalam@unsw.edu.au).

Bikram P. Banerjee is with the School of Surveying and Built Environment, University of Southern Queensland, Toowoomba, Queensland, 4350, Australia (e-mail: Bikram.Banerjee@unisq.edu.au).

Nancy. F. Glenn is with the Department of Geosciences, Boise State University, Boise, ID, USA (e-mail: nancyglenn@boisestate.edu).

Simit Raval is with the School of Minerals and Energy Resources Engineering, University of New South Wales, Sydney, Australia NSW 2052 (e-mail: simit@unsw.edu.au).

for user needs and assessing the quality of segmentation outputs is of significant value to the mining industry.

## II. MATERIALS AND METHODS

*A. Methodological overview*

The workflow employed for the comparative analysis of traditional segmentation methodologies, including mean shift segmentation, simple linear iterative clustering (SLIC), and Voronoi-based segmentation, alongside deep learning-based segmentation approaches such as Stardist segmentation and the Segment Anything Model (SAM), is given in Fig. 1. The colour-based segmentation techniques encompass mean shift segmentation and SLIC, while Voronoi-based segmentation, Stardist segmentation, and SAM are characterised as morphology-based segmentation methodologies.

The workflow commences with the creation of an orthomosaic and digital surface model (DSM), leveraging geotagged RGB images acquired through UAV data acquisition. Subsequent to this, the generation of ground truth segments ensues, providing a foundational reference for the subsequent evaluation of segmentation outcomes derived from both traditional and deep learning-based algorithms. Within the framework of traditional segmentation methodologies, the determination of optimal parameters is achieved through a comparative assessment of Hoover metric scores, thereby facilitating the attainment of segmentation outcomes of optimal quality. Simultaneously, fine-tuning of deep learning-based approaches is conducted to achieve precise spoil pile segmentation. In the case of deep learning methodologies, the hillshade derived from DSM was used as the input. This integrative approach ensures a refined and accurate segmentation process, contributing to the enhancement of both traditional and deep learning-based segmentation methodologies within the context of complex terrain analysis.

Subsequently, the performance evaluation is executed by comparing the selected optimal traditional segmentation under each respective algorithm with the outcomes obtained from fine-tuned deep learning algorithms. This comparative analysis is conducted utilising the Hoover metric, providing a quantitative assessment of segmentation effectiveness. The systematic workflow presented in Fig.1 ensures a structured evaluation of both traditional and deep learning-based segmentation methodologies in the context of UAV-derived products.

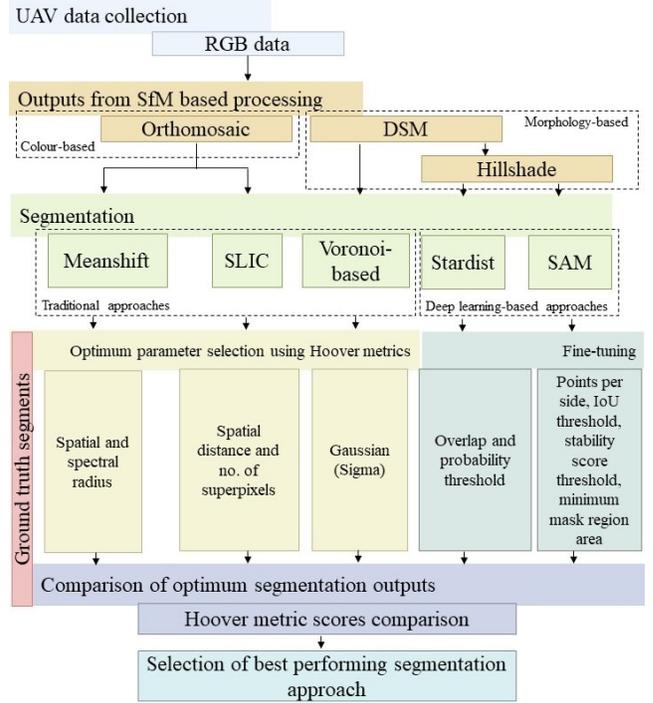

**Fig. 1.** Workflow employed for the comparative analysis of traditional and deep learning-based segmentation approaches in the context of coal spoil pile segmentation.

*B. Study site and data acquisition*

The study focused on a mining dump site employing paddock dumping in the Sydney basin, New South Wales, Australia (Fig. 1(e)). The data collection took place on 9th November 2022, using a DJI Matrice 300 RTK quadcopter equipped with a 45-megapixel RGB camera (Zenmuse P1 model) featuring a full-frame CMOS sensor (35.9 mm × 24 mm) and a 35 mm lens with a 63.5° field of view. The camera was mounted on a three-axis stabilised gimbal. The UAV system was configured to capture optical image data at solar noon, flying at a height of 80 m, resulting in a ground sampling distance (GSD) of 1.22 cm. The image acquisition settings included automated capture mode with 80% forward and side overlaps.

*C. Data processing and ground truth generation*

A Structure from Motion (SfM) based photogrammetric stitching package, specifically Pix4D Mapper (Pix4D SA, Lausanne, Switzerland), was employed for processing raw images acquired during the UAV mission. The SfM workflow encompassed several vital steps, initiating with image matching and subsequent tie points detection. Subsequently, external orientation was refined through bundle block adjustment. The ensuing phase focused on generating dense point clouds utilising high-quality parameters, facilitating the creation of point clouds characterised by a high density. Finally, the processing pipeline culminated in the generation of an orthomosaic (Fig. 2(a)), and a DSM (Fig. 2(b)) characterised by a spatial resolution of 5 cm. Furthermore, the hillshade representation (Fig. 2(c)) was derived by applying a sigmoidal stretch to the DSM with a specified strength level set at 3 and scaling factor of 2.

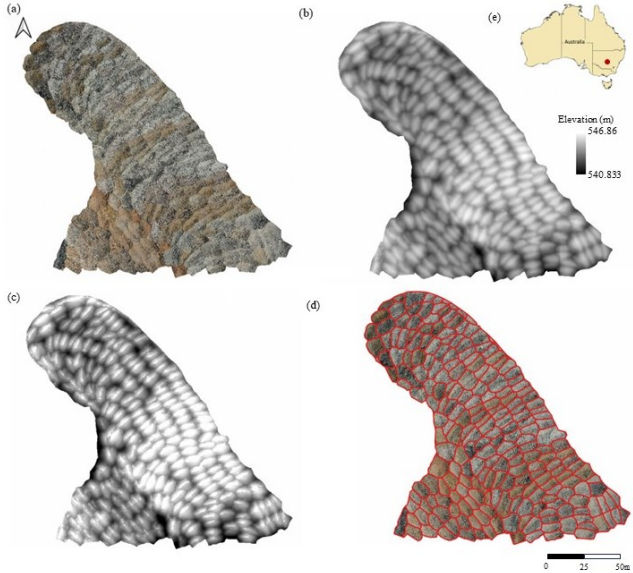

**Fig. 2. (a)** Orthomosaic and **(b)** digital surface model (DSM) and **(c)** hillshade representation of selected area of spoil piles, **(d)** ground truth segments and **(e)** location of selected area.

Ground truth segments were generated through a process of manual digitisation (Fig. 2(d)), focusing specifically on individual spoil piles within paddocks. This digitisation procedure was executed with reference to spatial data acquired through on-site field investigations and the visual interpretation of the orthomosaic. A total of two hundred thirty-two (232) ground truth segments were meticulously delineated within an area spanning 17174 $m^2$.

*D. Segmentation*

Object-based image classification comprises two sequential steps: orderly segmentation and subsequent classification. In the segmentation phase, pixels are systematically grouped into objects based on their homogeneity, which refers to the similarity of spectral values and spatial properties. This process involves the amalgamation of adjacent pixels that share common characteristics, facilitating the delineation of meaningful and coherent image objects. The objective is to create homogeneous segments that can be effectively employed for subsequent classification.

This study concentrated on investigating two distinct segmentation approaches: colour-based segmentation and morphological-based segmentation. Within the colour-based segmentation category, we explored two widely utilized iterative clustering algorithms, namely, the mean shift algorithm [9] and the SLIC algorithm [10]. Within the morphological-based approach, we considered a traditional method known as Voronoi-based segmentation [11]. Additionally, two deep learning-based segmentation approaches, namely, StarDist segmentation [12] and SAM [13], were also explored in an attempt to assess their efficacy in the context of spoil pile delineation. A brief description of algorithms related to this work is provided in the below sections.

1) **Mean shift segmentation**
Mean shift is a non-parametric method characterised by an iterative process that relocates each data point to the average position of data points within its neighbourhood [14]. This algorithm identifies dense areas within a feature space, commonly referred to as clusters or regions of local maxima. Through an iterative clustering process, mean shift effectively groups the data points. The essential parameters for the mean shift algorithm segmentation include the spatial radius, range radius, and minimal region size.

In this study, various combinations of these parameters were employed, and the segmentation output generated with optimal parameter settings was chosen for comparative analysis with other segmentation algorithms.

2) **Simple linear iterative clustering**
Superpixel algorithms aim to generate perceptually meaningful clusters from individual pixels. The SLIC algorithm, a representative superpixel algorithm, is built upon the foundation of the k-means algorithm [15]. Unlike the k-means method, which operates across the entire colour space, the SLIC algorithm confines its search region to a predefined area that encompasses the superpixel. This approach imparts certain advantages over the k-means method, including the capacity to control the compactness of superpixels. This is achieved by introducing a novel distance metric that considers both colour and spatial distances [16].

The SLIC approach generates superpixels by clustering pixels based on colour similarity and proximity in the 'labxy' five-dimensional (5D) space [15]. 'lab' represents the pixel colour vector in the perceptually uniform CIELAB colour space, and 'xy' denotes the spatial position. While colour distance is constrained in CIELAB, spatial distance in the 'xy' plane depends on image dimensions, requiring normalisation for Euclidean distance in the 5D space. SLIC employs a distance measure accounting for superpixel size to ensure uniform cluster sizes and spatial extents, aiming for balanced colour and spatial influence in clustering for consistent superpixel creation.

In the context of this study, the optimisation of the SLIC algorithm involved considering parameters such as spatial distance and the number of superpixels. These parameters were systematically adjusted to achieve optimal segmentation results, aligning with the specific objectives and requirements of the study.

3) **Voronoi-based segmentation**
The Voronoi-based segmentation approach [17] encompasses four steps: noise reduction, seed point detection, background seed point removal, and pile polygonisation based on Voronoi tessellation. The methodology initiates with the application of Gaussian blurring to the image for noise reduction, followed by the detection of local maxima. These steps contribute to the identification of seed points crucial for subsequent segmentation. Subsequently, Otsu's thresholding method [18] is employed to differentiate between the background and pile foregrounds. Background seed points are then eliminated through a binary operation involving the detected seed points and the background. The final step involves drawing polygonal areas corresponding to the piles based on Voronoi tessellation computed from the identified seed points. In this study, the sigma values of the Gaussian blur were systematically adjusted to optimise the algorithm for the precise detection of piles.

4) **StarDist segmentation**

The StarDist model [19] employs a methodology wherein it anticipates a star-convex polygon for each pixel within the image. Precisely, it engages in the regression of distances from the pixel to the boundary of the corresponding object, considering a predetermined set of n radial directions characterised by equidistant angles. This process is applicable exclusively to pixels within an object, excluding those associated with the background. Consequently, the StarDist model undertakes the discrete prediction of the likelihood of a pixel being part of an object, ensuring that polygons are derived solely from pixels with sufficiently elevated object probability. Following the generation of polygon candidates and their respective object probabilities, a non-maximum suppression (NMS) technique is implemented to yield the ultimate set of polygons, each representing a distinct object instance.

In contrast to a binary mask-based classification approach designating pixels as either object or background, the StarDist model adopts an alternative strategy. Specifically, the object probability of each pixel is defined as the normalized Euclidean distance to the nearest background pixel. This distinctive approach is strategically chosen to optimize the performance of the subsequent non-maximum suppression process. By ascribing object probabilities based on the Euclidean distance metric, NMS tends to prioritize polygons associated with pixels situated closer to the center of the object, a characteristic indicative of more accurate object representation.

Regarding the computation of distances for pixels within an object, the StarDist model calculates the Euclidean distances to the object boundary by traversing along each radial direction until encountering a pixel with a distinct object identity. This methodology ensures a comprehensive determination of distances for pixels constituting an object, facilitating subsequent analysis and predictions within the model's framework.

The StarDist algorithm is underpinned by the U-Net architecture. Specifically, it adopts the foundational structure of U-Net while incorporating a refinement in the form of an additional convolutional layer subsequent to the final U-Net feature layer. This strategic inclusion serves to mitigate potential conflicts between the two subsequent output layers, thereby enhancing the algorithm's efficacy. Notably, the ultimate layer in the architecture is responsible for generating the object probability. The implementation of StarDist is encapsulated within a Python package, complete with pre-trained models tailored for 2D applications. Of particular relevance is the utilisation of a model specifically trained on nuclei images, which aptly captures the morphological characteristics inherent in the hillshade dataset representing spoil piles central to the investigative focus of this work. The rationale behind employing this model lies in the discernible congruence between the morphological patterns observed in nuclei images and the hillshade depictions of the spoil piles.

This study utilised the hillshade representation of the DSM as input for StarDist segmentation, employing probability and overlap thresholds of 0.3 and 0.9 for optimal segmentation. Systematic adjustments of these parameters were implemented by varying the thresholds in different combinations. Subsequent to these adjustments, outcomes were subjected to visual assessment to ascertain the accuracy of the segmentation process.

5) **Segment anything model (SAM) based segmentation**

Meta AI has recently introduced a revolutionary foundation model called the SAM [13]. Foundation models are deep neural networks trained using self-supervision and transfer learning on massive datasets, enabling them to adapt to various tasks and serve as the basis for diverse applications beyond those encountered during training. SAM represents a significant advancement in image segmentation methodologies. The SAM model's architecture comprises three core components: an image encoder, a prompt encoder, and a mask decoder. The image encoder processes the input image, while the prompt encoder handles both sparse and dense prompts. The mask decoder generates a segmentation mask. This architecture allows SAM to predict multiple masks for the same instance.

In this study, SAM was applied to a hillshade representation of the spoil pile segmentation. A grid of points (200 points per side) was overlaid on the scene, and SAM generated all predicted masks, segmenting the entire scene at once. The study employed an intersection of union (IoU) threshold and stability score threshold of 0.9 and 0.9, respectively. Additionally, the number of crops, wherein an image crop denotes a smaller region extracted from a larger image to facilitate localised processing, was fixed at 1. The minimum mask region area, which was kept at 100 for optimal segmentation. These parameters were systematically adjusted, and the results were visually assessed to ensure accurate segmentation.

*E. Performance evaluation of segmentations*

This study investigated the optimal parameter settings for traditional segmentation techniques and evaluated their performance against alternative algorithms. Hoover metrics [20], which quantifies the discrepancies between ground truth and machine-generated segmentations, were employed to assess the segmentation accuracy in terms of perfect segmentation, over-segmentation (a single object being partitioned into multiple objects), under-segmentation (inadequate partitioning of multiple objects), and missed detection (failure to detect an object). The segmentation evaluation was conducted using an overlapping threshold of 0.5. Four performance metrics, namely, correct detection score, over-segmentation score, under-segmentation score, and missed detection, were calculated and compared to comprehensively evaluate the segmentation performance of the algorithms. A higher score for each metric serves as a determinant in ascertaining whether the segmentation algorithm yields accurate segmentation, over-segmentation, under-segmentation, or missed segmentation.

### III. RESULTS AND DISCUSSION

*A. Optimal parameter selection for traditional segmentation approaches*

The Hoover metrics scores for mean shift segmentation, SLIC segmentation, and Voronoi-based segmentation, along with their corresponding segmentation parameters, are presented in Figures 3, 4, and 5, respectively.

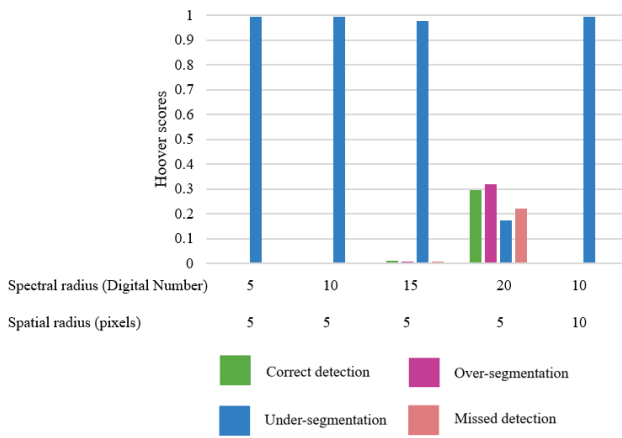

**Fig. 3.** Hoover metric scores corresponding to parameters in mean shift segmentation. Optimal performance, indicative of accurate detection, is attained with a spectral radius of 20 digital numbers and a spatial radius of 5 pixels. Alternative parameter combinations yielded increased level of under-segmentation.

Mean shift segmentation exhibited improved segmentation performance when employing high spectral radii and low spatial radii at minimum region size of 10,000 pixels. The spectral radius serves as a parameter denoting the relative importance in discerning spoils based on their colour characteristics. A lower spectral radius is considered appropriate for distinguishing features characterised by closely aligned spectral attributes. Conversely, the spatial radius is a parameter reflecting the relative emphasis placed on segregating spoils according to their spatial characteristics. A lower spatial radius is deemed suitable for discriminating features that are both spatially confined and closely clustered. This finding (Fig. 3) can be attributed to the need for substantial smoothing to eliminate the fine granularity of individual piles, necessitating a low spatial radius as the optimal segmentation parameter. Conversely, the high spectral radius is attributable to the requirement for algorithms to differentiate between piles with similar spectral characteristics. While mean shift segmentation achieved the highest correct detection scores at a spectral radius of 20 DN and a spatial radius of 5 pixels, it is noteworthy that over-segmentation was also high (0.319) with these parameter settings. This indicates that even though the segmentation using algorithm does not precisely match the ground truth segmentation, it can still contribute to accurate object-based classification. This is because over-segmented polygons also fall within the geometry of ground truth segments and share their material properties. Therefore, both correctly detected and over-segmented piles, culminating in a Hoover score of 0.614, contribute to accurate classification.

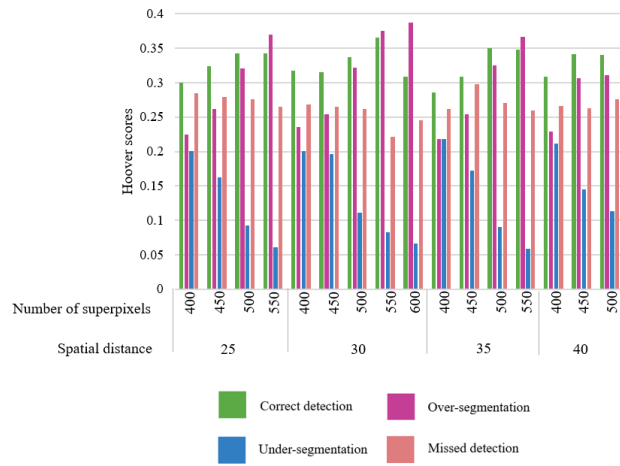

**Fig. 4.** Hoover metrics scores for specific parameters in Simple Linear Iterative Clustering (SLIC). The correct detection score is maximised at spatial distances of 30 and 550 superpixels. However, at these parameter settings, the oversegmentation score surpasses the correct detection score. Increasing the number of superpixels while maintaining a spatial distance of 30 leads to a rise in oversegmentation, accompanied by a decline in correct detection.

The SLIC algorithm's performance (Fig. 4) exhibits a peak correct detection score of 0.365 at spatial distances of 30 and 550 superpixels. In the context of large superpixels, spatial distances take precedence over colour proximity, thereby affording greater relative significance to spatial proximity than to colour. This phenomenon results in the generation of compact superpixels that exhibit limited adherence to image boundaries. Conversely, for smaller superpixels, colour proximity supersedes spatial distances, assigning greater relative importance to colour proximity than to spatial proximity [15]. In this specific instance, a spatial distance of 30 and 550 superpixels has been identified as an appropriate parameter, facilitating the delineation of more discernible spoil piles. However, the substantial heterogeneity within the piles leads to over-segmentation of most piles, yielding an oversegmentation score of 0.375. This over-segmentation stems from SLIC's pixel clustering approach, which utilises both spectral similarity and proximity within piles. Consequently, the aggregate contribution of correctly and oversegmented piles to accurate classification amounts to 0.74 in Hoover scores.

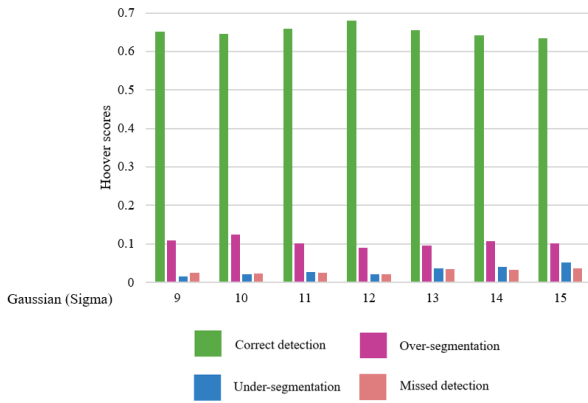

**Fig. 5.** Hoover metrics scores for corresponding parameters of Voronoi-based segmentation. The correct detection scores are presented in relation to the sigma values of Gaussian blurring. Notably, a superior correct detection score is observed at a sigma value of 12.

The Voronoi-based segmentation algorithm's performance (Fig. 5) is maximised when the sigma value of Gaussian blurring is set to 12. This optimal performance is attributed to the balanced noise removal achieved at this sigma value. Sigma values lower than 12 fail to adequately eliminate noise, hindering the identification of individual pile maxima. Conversely, sigma values higher than 12 blur the images to an extent that distinct piles are no longer distinguishable for local maxima detection. Notably, both correctly and over-segmented piles at sigma value of 12 exhibit a Hoover score of 0.77.

*B. Optimum segmentation using deep learning-based approaches*

1) **StarDist Segmentation**

StarDist identifies individual piles by predicting the distances to object boundaries along a fixed set of rays (Fig. 6(a)), resulting in a set of polygons that correspond to piles in the input image, as shown in Fig. 6(b).

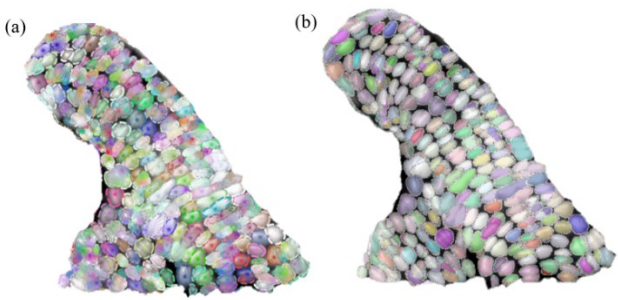

**Fig. 6. (a)** Star-convex polygons parameterised by the radial distances in hillshade representation of spoil pile and **(b)** StarDist segmentation results.

StarDist achieved a Hoover correct detection score of 0.494 and an over-segmentation score of 0.242. Both correctly detected, and over-segmented piles contributed to an overall accuracy of 0.736. These scores suggest that StarDist is more sensitive to noise and that the piles in the dataset are densely packed. Additionally, the internal structure of the piles can cause over-segmentation. For example, StarDist can split elongated cells into multiple star-convex subsets that do not span the entire pile [21]. The StarDist network can make robust predictions of its distance field, but it fails to assemble accurate pile masks in some cases because some piles in the dataset are not well-approximated by star-convex polygons.

2) **SAM based Segmentation**

Fig. 7 presents the overlayed segmentation generated using SAM. SAM achieved a remarkable Hoover correct detection score of 0.714 and an over-segmentation score of 0.226. The combined contribution of accurately detected and over-segmented piles resulted in an exceptional overall accuracy of 0.940. Zero-shot learning empowers a model to effectively process and respond to input data that it has not explicitly encountered during the training phase. This remarkable ability stems from the model's acquisition of a generalised understanding of the data rather than a mere collection of specific instances, culminating in accurate segmentation [13]. Zero-shot learning systems exhibit the remarkable ability to recognise piles they have never encountered before, relying on their understanding of underlying concepts and relationships.

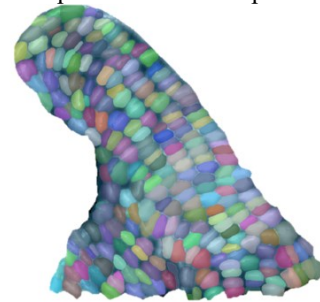

**Fig. 7.** Segmented piles using segment anything model (SAM)

*C. Comparison of traditional and deep learning-based segmentation approaches*

The performance of pile segmentation algorithms can be assessed through the Hoover score, a comprehensive metric that considers correct Detection, over-segmentation, under-segmentation, and missed detection (Fig. 8).

Among the algorithms evaluated, mean shift demonstrates a relatively balanced performance, with moderate scores across all metrics. SLIC exhibits a higher correct detection rate, indicating a better ability to accurately identify segments, but at the cost of increased over-segmentation. Voronoi, on the other hand, achieves an impressive correct detection score with minimal over-segmentation and under-segmentation, suggesting robust performance in delineating objects in the image. StarDist strikes a balance between correct detection and over-segmentation, showcasing a competitive overall performance. Notably, SAM stands out with the highest correct detection score, implying superior accuracy.

Upon closer examination, SAM appears to be particularly well-suited for spoil pile segmentation where precision is paramount. Its high correct detection score indicates a strong capability to identify segments accurately, while simultaneously minimising errors.

The selection of an image segmentation algorithm is contingent upon the specific requirements of the given task. Both SAM and Voronoi demonstrate promising outcomes in terms of overall performance. SAM exhibits notable strength, whereas Voronoi excels in achieving closer and more accurate detections.

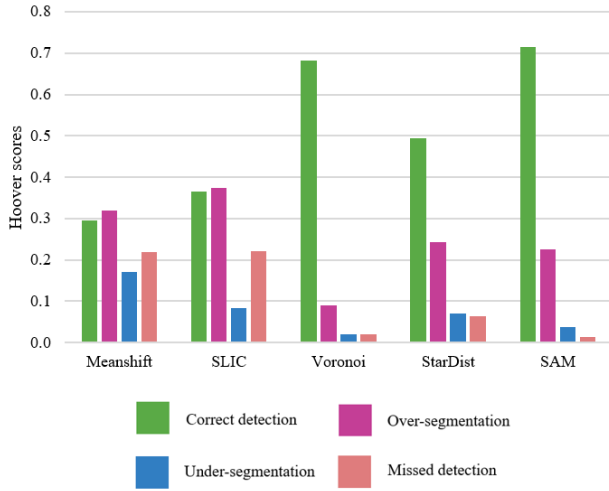

**Fig. 8.** Hoover metric scores for best performing traditional and deep learning-based segmentation approaches

Fig. 9 and 10 provide a visual representation of segmentation performance. Visual assessment of segments from Fig. 9 and 10 align with the scores derived from Hoover metrics in Fig. 8.

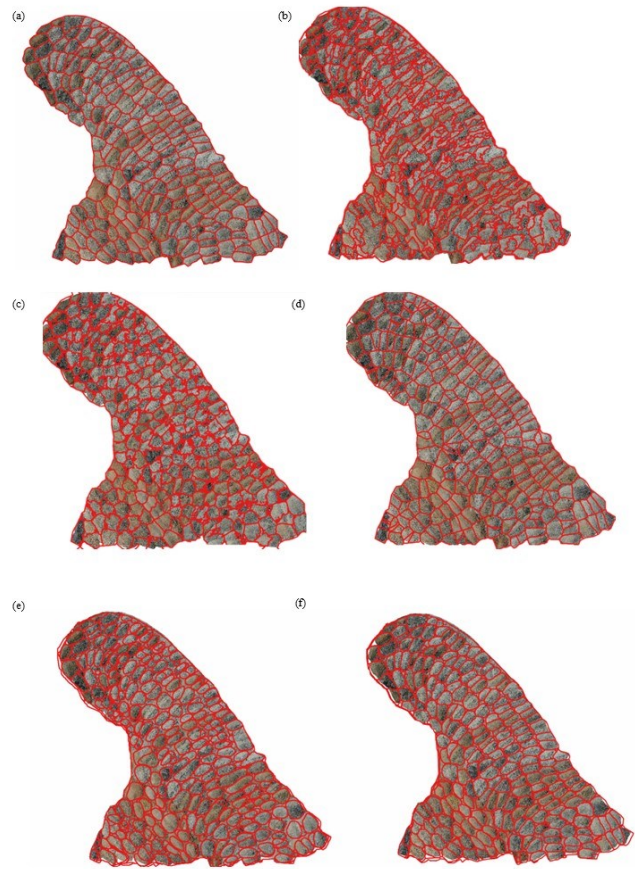

**Fig. 9. (a)** Ground truth segments and segmentation results of **(b)** mean shift algorithm, **(c)** simple linear iterative clustering, **(d)** Voronoi-based segmentation, **(e)** StarDist segmentation and **(f)** segment anything model

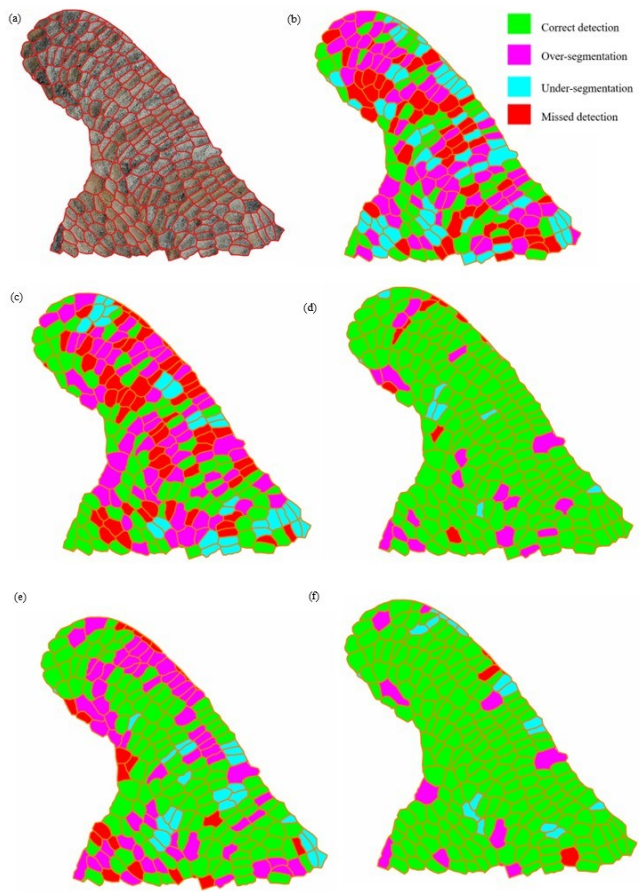

**Fig. 10. (a)** Ground truth segments and Hoover scores on ground truth for (b) mean shift algorithm, (c) simple linear iterative clustering and (d) Voronoi-based segmentation, (e) StarDist segmentation and (f) segment anything model

The assessment of segmentation methodologies indicates that morphology-based approaches (specifically Voronoi, StarDist, and SAM-based segmentation) outperform colour-based methods (mean shift and SLIC) in accurately delineating paddock tipped dumps. This superiority may stem from the susceptibility of colour-based approaches to external factors, such as shadows and variations in sun angles during image acquisition with UAVs. These factors can lead to the misclassification of pixels from the same class into distinct categories. Notably, morphology-based segmentation, incorporating DSM and hillshade representations, remains less influenced by these external factors.

Among the evaluated segmentation approaches, Voronoi-based segmentation and SAM-based segmentation exhibit the highest correct detection rates at 0.681 and 0.714, respectively. It is noteworthy that the combination of correct detection and over-segmentation scores, where over-segmented polygons align with the geometry of ground truth segments and share their material properties, results in values of 0.771 and 0.940, respectively. This underscores the effectiveness of deep learning-based segmentation, with a particular emphasis on the performance of SAM in this context. This underscores the superior performance of deep learning approaches over traditional methods, showcasing their zero-shot learning capability.

## IV. CONCLUSIONS

The mining industry is undergoing a paradigm shift towards image-based analysis, driven by the imperative to alleviate the labour-intensive and inherently hazardous visual-based characterisation process. This study, centered on object-based image analysis, heralds a new era for the industry, paving the way for enhanced efficiency in spoil pile characterisation. However, the efficacy of advanced image processing techniques, such as object-based classification and feature extraction, hinges critically on segmentation quality. Consequently, identifying suitable spoil pile delineation algorithms becomes a pivotal step prior to classification. This study examines different ways to separate spoil piles, focusing on colour and morphology methods. The results indicate that using morphology-based segmentation outperforms colour-based methods for accurately defining spoil piles. Colour-based approaches exhibit a pronounced reliance on optical data, which is susceptible to perturbations induced by external factors such as sun angle and shadows in the undulating terrain of spoil piles, leading to misclassification of classes. In contrast, the morphology-based approach utilises DSM, which remains immune to these external illumination factors. Additionally, a comparison of traditional segmentation approaches with deep learning-based approaches underscores the effectiveness of deep learning-based approaches, particularly SAM, in the spoil pile delineation task. This efficacy stems from their zero-shot learning capability and inherent adaptability. Moreover, this investigation systematically explores the nuances associated with the selection of optimal parameters for segmentation methodologies applied to the delineation of spoil piles. The overarching objective is to augment the quality of segmentation outcomes. In the specific instance of SAM, four parameters were meticulously fine-tuned to achieve precision in spoil pile delineation, a process contingent upon the spatial resolution of the DSM. It is imperative to underscore that this fine-tuning approach holds transferability, as it can be extrapolated to other sites with paddock dumping practices. Moreover, adapting and fine-tuning these parameters for varying spatial resolutions is anticipated to contribute significantly to the enhancement of segmentation efficacy across different contexts. This study provides invaluable insights into segmentation approaches and corresponding parameters for spoil pile delineation, laying the foundation for automated image-based spoil characterisation in the impending future.